\documentclass{article}

\PassOptionsToPackage{numbers, compress}{natbib}

     \usepackage[preprint]{neurips_2020}

\usepackage[utf8]{inputenc} 
\usepackage[T1]{fontenc}    
\usepackage{hyperref}       
\usepackage{url}            
\usepackage{booktabs}       
\usepackage{amsfonts}       
\usepackage{nicefrac}       
\usepackage{microtype}      
\usepackage{amsmath}
\usepackage{subfigure}
\usepackage{graphicx}
\usepackage{bm}
\usepackage{amsthm}
\usepackage{xcolor}
\usepackage{multirow}
\usepackage[ruled,vlined]{algorithm2e}
\usepackage{colortbl}
\usepackage{wrapfig}
\usepackage[toc,page]{appendix}
\definecolor{mygray}{gray}{.9}

\SetCommentSty{mycommfont}
\SetKwInput{KwInput}{input}                
\SetKwInput{KwOutput}{output}              

\newtheorem{remark}{Remark}

\usepackage{subfigure}
\usepackage{subcaption}
\usepackage{float}

\title{CF-GO-Net: A Universal Distribution Learner via Characteristic Function Networks with Graph Optimizers}

\author{%
    Zeyang Yu \\
	Department of Electrical and Electronic Engineering \\
	Imperial College London \\
	\texttt{z.yu17@imperial.ac.uk} \\
    \AND
	Shengxi Li\\
	Department of Electrical and Electronic Engineering\\
	Imperial College London\\
	\texttt{shengxi.li17@imperial.ac.uk} \\
	\AND
	Danilo Mandic \\
	Department of Electrical and Electronic Engineering \\
	Imperial College London \\
	\texttt{d.mandic@imperial.ac.uk} \\
}

\begin{document}
\maketitle
\begin{abstract}
	Generative models aim to learn the distribution of datasets, such as images, so as to be able to generate samples that statistically resemble real data. However, learning the underlying probability distribution can be very challenging and intractable. To this end, we introduce an approach which employs the characteristic function (CF), a probabilistic descriptor that directly corresponds to the distribution. However, unlike the probability density function (pdf), the characteristic function not only always exists, but also provides an additional degree of freedom, hence enhances flexibility in learning distributions. This removes the critical dependence on pdf-based assumptions, which limit the applicability of traditional methods. While several works have attempted to use CF in generative modeling, they often impose strong constraints on the training process. In contrast, our approach calculates the distance between query points in the CF domain, which is an unconstrained and well defined problem. Next, to deal with the sampling strategy, which is crucial to model performance, we propose a graph neural network (GNN)-based optimizer for the sampling process, which identifies regions where the difference between CFs is most significant. In addition, our method allows the use of a pre-trained model, such as a well-trained autoencoder, and is capable of learning directly in its feature space, without modifying its parameters. This offers a flexible and robust approach to generative modeling, not only provides broader applicability and improved performance, but also equips any latent space world with the ability to become a generative model.

\end{abstract}

\section{Introduction}

Generative Adversarial Networks (GANs) have become de-facto standard in generative modeling \cite{goodfellow2014generative}, whereby two neural networks—a generator and a discriminator—are trained together. The generator creates data samples which aim to mimic the real data, while the discriminator tries to distinguish between the real and generated data. As training progresses, the generator improves at producing more realistic data, making it more difficult for the discriminator to distinguish between the two.

Despite their success, GANs face several key challenges, with some of the main issues including \textbf{mode collapse}, whereby the generator produces only a limited variety of samples; \textbf{gradient vanishing}, where updates to the generator become too small to make progress; and \textbf{unstable training}, characterized by the generator and discriminator either failing to converge or exhibiting oscillatory behaviour \cite{salimans2016improved}. These challenges have restricted the potential of GANs, prompting the development of more stable and robust generative models.

One such improvement is the \textbf{f-GAN}, which extends adversarial training through f-divergence, a statistical measure that quantifies the difference between two probability distributions \cite{nowozin2016f}. The f-divergence between two distributions \( P \) and \( Q \) is defined as

\begin{equation}
    D_f(P \parallel Q) = \int_{\mathcal{X}} q(x) f\left(\frac{p(x)}{q(x)}\right) dx
    \label{eq:divergence}
\end{equation}

where \( p(x) \) and \( q(x) \) are the respective densities of the distributions \( P \) and \( Q \), and \( f \) is a convex function. While this provides more flexibility than the Jensen-Shannon divergence used in the original GAN, it is still difficult to directly minimize such a divergence measure.

To address this issue, f-GANs use the \textbf{convex conjugate} of \( f \), given by

\begin{equation}
    f^*(t) = \sup_{u \in \text{dom}_f} \left( ut - f(u) \right)
    \label{eq:conjugate}
\end{equation}
which simplifies the problem by turning it into a dual form that can be solved efficiently and allows the f-divergence to be expressed in a tractable form. Furthermore, the \textbf{Legendre transform} can be used to further simplify \eqref{eq:divergence} into the maximization over a discriminator function \( T \), focusing on the first-order derivative of \( f \) in the form

\begin{equation}
    D_f(P \parallel Q) = \sup_{T \in \text{dom}(f')} \mathbb{E}_{x \sim P} \left[T(x)\right] - \mathbb{E}_{x \sim Q} \left[f^*(T(x))\right]
    \label{eq:f-divergence-sup}
\end{equation}

In this framework, \( T \) serves as the discriminator, with the form of \( f^* \) determined by the choice of the f-divergence. By designing custom loss functions, this flexibility makes f-GANs adaptable to various tasks.

Another important class of GANs is the \textbf{Integral Probability Metric (IPM) GAN}, which optimizes the distance between real and generated distributions by relying on specific function spaces \cite{muller1997integral}. The general form of an IPM between two distributions \( P \) and \( Q \) is given by the "worst case" form

\begin{equation}
    \text{IPM}_{\mathcal{F}}(P, Q) = \sup_{f \in \mathcal{F}} \left( \mathbb{E}_{x \sim P} \left[f(x)\right] - \mathbb{E}_{x \sim Q} \left[f(x)\right] \right)
    \label{eq:ipm}
\end{equation}

where \( \mathcal{F} \) is a function class that defines the metric. The discriminator (or \textit{critic}) searches within this function space with the aim to maximize the difference between the two expectations. For example, the \textbf{Wasserstein GAN (WGAN)} employs the Earth Mover's Distance (EMD), also called the Wasserstein-1 distance, to minimize the cost of transporting probability mass from one distribution to another \cite{arjovsky2017wasserstein}. This leads to more stable training and better convergence. Other variants of IPM-GANs extend this approach by employing different metrics, such as the \textbf{Fisher GAN}, which uses a second-order constraint to link the optimization problem to the Chi-squared distance \cite{mroueh2017fisher}, and \textbf{MMD GAN}, which uses kernel methods to define the Maximum Mean Discrepancy (MMD) \cite{li2017mmd}.

It is important to recall that GANs were originally framed as an adversarial game, where the generator tries to fool the discriminator while the discriminator seeks to distinguish real from generated data. However, much of the subsequent development in GANs has focused on designing and optimizing more advanced distance metrics between distributions. Models such as f-GANs and IPM GANs have evolved to approximate and minimize these metrics using advanced optimization techniques. These models often employ dual formulations, such as the convex conjugate \eqref{eq:conjugate} in f-GANs or the Kantorovich-Rubinstein duality in WGANs, to provide better theoretical guarantees and improve training stability. This shift toward distance metric design has helped address many challenges inherent to the original adversarial framework, such as mode collapse and unstable gradients, by aligning the optimization with more robust and effective distance measures that capture the geometry of the data \cite{ansari2020characteristic, arjovsky2017wasserstein}.

To broaden the potential applications of generative modeling across both a range of probabilistic frameworks and latent space models which so far have not been able to generate data, we propose a novel approach based on minimizing the distance between characteristic functions (CFs). While several previous works have attempted to incorporate CFs into generative models, they often impose severe constraints on how the model is trained \cite{ansari2020characteristic}. Our method, in contrast, calculates the distance between query points in the CF domain, while the innovative sampling strategy plays a crucial role in model performance. This is achieved by employing a graph neural network (GNN)-based strategy which optimises this sampling by identifying regions where the difference between CFs is most significant \cite{Stankovi2020GraphSP}. The proposed approach is justified by the Levy continuity theorem, which guarantees that the empirical characteristic function (ECF) converges to the true CF as the sample size increases, thus ensuring stability. The graph optimizer enables efficient and optimized evaluation of the distance between characteristic functions, leads to improved performance and offers three key advantages:
\begin{itemize}
    \item \textbf{Efficient distribution measurement}: Working in the CF domain provides a stable, natural measure of distributional distance, thus simplifying the training process by directly comparing entire distributions rather than individual data samples. In addition, such an approach measures the overall distance between batches, giving a clearer view of distributional discrepancies.
    
    \item \textbf{Dynamic sampling with GNNs}: The GNN-based optimization dynamically adapts the sampled query points to focus on regions where discrepancies between the real and generated data are most pronounced, especially in complex feature spaces. Without the use of GNNs, the CF-based approach would have been limited to only simpler distributions. However, the GNN extends the ability of CFs to handle more intricate data distributions.

    \item \textbf{Flexible adaptation of standard pre-trained non-generative models}: Our approach allows for well-trained non-generative models—such as autoencoders, compression models, or denoising models—to be transformed into generative models without requiring re-training or fine-tuning. By freezing the parameters of pre-existing models, we optimize the generator to match the target distribution in its feature space. This flexibility enables full use of models which are already effective in  tasks such as feature extraction and expands their application to generative tasks.
\end{itemize}

Overall, our proposed method opens new avenues for generative modeling by enabling the transformation of non-generative models into high-performance generative systems, particularly for tasks requiring complex feature representations.

\section{Characteristic Function Loss and Graph Network Optimizer}
We shall now revisit the characteristic function in view of comparing two distributions.

\subsection{Characteristic Function Loss}

The characteristic function (CF) of a random variable $\bm{\mathcal{X}} \in \mathbb{R}^m$ is the statistical expectation of its complex exponential transform, given by
\begin{equation}\label{eq_chf}
\Phi_{\bm{\mathcal{X}}}(\mathbf{t}) = \mathbb{E}_{\bm{\mathcal{X}}}[e^{j\mathbf{t}^T\mathbf{x}}] = \int_\mathbf{x} e^{j\mathbf{t}^T\mathbf{x}}dF_{\bm{\mathcal{X}}}(\mathbf{x}),
\end{equation}
where $F_{\bm{\mathcal{X}}}(\mathbf{x})$ is the cumulative distribution function (CDF) of $\bm{\mathcal{X}}$. Unlike the probability density function (PDF), which may not exist for all distributions, CFs are both guaranteed to exist and retain all the information about the distribution. Additionally, CFs are automatically aligned at $\mathbf{t} = \mathbf{0}$, thus simplifying comparisons between distributions.

Since the true CF is typically unknown, we rely on the Empirical Characteristic Function (ECF), which approximates the CF using sample data. Given a set of $n$ independent and identically distributed (i.i.d.) samples $\{\mathbf{x}_i\}_{i=1}^n$ from $\bm{\mathcal{X}}$, the ECF is defined as
\[
\widehat{\Phi}_{\bm{\mathcal{X}}_n}(\mathbf{t}) = \frac{1}{n} \sum_{i=1}^n e^{j\mathbf{t}^T \mathbf{x}_i}.
\]

The distinguishing properties of CFs for its application in generative models are:
\begin{itemize}
    \item The Levy continuity theorem \cite{williams1991probability} ensures that as the sample size increases, the ECF converges to the true CF. 
    
    \item Additionally, the uniqueness theorem \cite{lukacs1972survey} guarantees that two distributions are identical if and only if their CFs are the same, making the ECF a powerful tool for comparing distributions, even when only finite samples are available.

    \item Another key property of the CF is that its higher-order derivatives at zero correspond to the higher-order moments of the data distribution. Specifically, the \(n\)-th order derivative of the CF at $\mathbf{t} = \mathbf{0}$ is related to the \(n\)-th order moment of $\bm{\mathcal{X}}$. 
\end{itemize}
Therefore, by focusing on values of the CF near zero, our method leverages the convergence properties, uniqueness, and higher-order moment information of the CF, making this region particularly important for accurately representing the statistical properties of data

To measure the distance between two random variables $\bm{\mathcal{X}}$ and $\bm{\mathcal{Y}}$, we define the CF loss as

\begin{equation}\label{eq_chfloss}
\mathcal{C}_{\bm{\mathcal{T}}}(\bm{\mathcal{X}}, \bm{\mathcal{Y}}) = \int_\mathbf{t} \big( ({\Phi}_{\bm{\mathcal{X}}}(\mathbf{t}) - {\Phi}_{\bm{\mathcal{Y}}}(\mathbf{t}))({\Phi}^*_{\bm{\mathcal{X}}}(\mathbf{t}) - {\Phi}^*_{\bm{\mathcal{Y}}}(\mathbf{t})) \big)^{\nicefrac{1}{2}} dF_{\bm{\mathcal{T}}}(\mathbf{t}),
\end{equation}

where ${\Phi}^*$ denotes the complex conjugate of ${\Phi}$, and $F_{\bm{\mathcal{T}}}(\mathbf{t})$ is the CDF of a sampling distribution on $\mathbf{t}$. The quadratic term for each $\mathbf{t}$ is represented as

\begin{equation}\label{eq_echfloss}
c(\mathbf{t}) = ({\Phi}_{\bm{\mathcal{X}}}(\mathbf{t}) - {\Phi}_{\bm{\mathcal{Y}}}(\mathbf{t}))({\Phi}^*_{\bm{\mathcal{X}}}(\mathbf{t}) - {\Phi}^*_{\bm{\mathcal{Y}}}(\mathbf{t})).
\end{equation}

\begin{remark}
The distinguishing property of $\mathcal{C}_{\bm{\mathcal{T}}}(\bm{\mathcal{X}}, \bm{\mathcal{Y}})$ is that it is a valid distance metric that measures the discrepancy between the CFs of two random variables, and it equals zero if and only if the two distributions are identical.
\end{remark}

Furthermore, the CF loss is complex-valued, which adds one more degree of freedom in the design. It can be decomposed into physically meaningful terms representing amplitude and phase differences \cite{yu2019widely, 9730054} to give

\begin{equation}\label{eq_chfloss_deomposed}
c(\mathbf{t}) = \underbrace{(|{\Phi}_{\bm{\mathcal{X}}}(\mathbf{t})| - |{\Phi}_{\bm{\mathcal{Y}}}(\mathbf{t})|)^2}_{\text{amplitude difference}} + 2|{\Phi}_{\bm{\mathcal{X}}}(\mathbf{t})||{\Phi}_{\bm{\mathcal{Y}}}(\mathbf{t})| \underbrace{(1 - \cos(\mathbf{a}_{\bm{\mathcal{X}}}(\mathbf{t}) - \mathbf{a}_{\bm{\mathcal{Y}}}(\mathbf{t})))}_{\text{phase difference}},
\end{equation}

where $\mathbf{a}_{\bm{\mathcal{X}}}(\mathbf{t})$ and $\mathbf{a}_{\bm{\mathcal{Y}}}(\mathbf{t})$ represent the respective angles (phases) of ${\Phi}_{\bm{\mathcal{X}}}(\mathbf{t})$ and ${\Phi}_{\bm{\mathcal{Y}}}(\mathbf{t})$. The amplitude in \eqref{eq_chfloss_deomposed} represents the diversity of the distribution while the phase reflects the data center. In our previous work \cite{yu2019widely}, we demonstrated that the amplitude and phase differences have varying importance depending on the task in hand, which is one of the key advantages of splitting the CF loss into its components.

Building on this decomposition, in our recent Reciprocal GAN work \cite{9730054}, we applied a joint training approach for the encoder and decoder under the assumption that the feature space followed a reciprocal relationship between the generator and the discriminator. 

\begin{remark}
This reciprocal relationship with the Reciprocal GAN enforces a Gaussian-like constraint on the feature space. For such a Gaussian-constrained feature space, a straightforward sampling strategy, like Gaussian sampling for the loss surface, is sufficient. However, Gaussian sampling strategy is limited to feature spaces with Gaussian-like structures. 
\end{remark}

Therefore, in more complex scenarios, such as those involving well-trained autoencoders with intricate feature spaces, a more sophisticated sampling approach is necessary to effectively capture the underlying distributional characteristics.

\subsection{Graph Neural Network as a Worst-Case Optimizer}

To address the challenges posed by complex feature spaces and the problem of ill-posed optima in sampling from $F_{\bm{\mathcal{T}}}(\mathbf{t})$, we introduce a Graph Neural Network (GNN) as a smart sampler that acts as a worst-case optimizer \cite{Stankovi2020GraphSP}. As discussed in our previous work \cite{9730054}, directly optimizing $F_{\bm{\mathcal{T}}}(\mathbf{t})$ can lead to instability, where it converges to point mass distributions, thus reducing the stability and the ability of the CF loss to distinguish between $\bm{\mathcal{X}}$ and $\bm{\mathcal{Y}}$. By leveraging the locality of GNNs, we are able to avoid this issue as the GNN dynamically updates sampling points based on local information \cite{Stankovi2019GraphSP_1}. This ensures that optimization remains stable and effective, even when dealing with complex feature spaces that are difficult to learn, especially in cases where the model is trained without strong regularization, yielding feature spaces over which we have little control\cite{ Stankovi2019GraphSP_2}.

Another key advantage of GNNs lies in their ability to optimize a dynamic loss surface while preserving local information during optimization. Each time the generator updates, the Empirical Characteristic Function (ECF) of the generated data shifts, altering the loss surface. Fully connected networks struggle in this context because they operate on a fixed mapping between input and output, making the original sampling points suboptimal after generator updates. On the other hand, GNNs are able to incorporate local information from the evolving loss surface and dynamically adapt the sampling points as the generator updates. By continuously adjusting to the changes in the loss surface, GNNs ensure that the worst-case regions are always prioritized during optimization. 

\begin{remark}
    This adaptability makes GNNs particularly suited for minimizing the discrepancies between the real and generated data distributions, even through these distributions are evolving during training. This dynamic adaptation prevents sampling from collapsing to suboptimal configurations, a common issue with methods relying on fixed distributions.
\end{remark}

We next introduce the GNN-based approach in more detail and provide experimental validation to demonstrate its effectiveness in optimizing the sampling process.

\section{Graph Neural Network as a Distribution Optimizer}

Building on the discussion of the advantages of GNNs in handling data locality and dynamically optimizing the loss surface, we now detail how the GNN is implemented as a distribution optimizer. The primary goal of our graph neural network (GNN) is to transform a set of random Gaussian sample points into new points that can identify the worst-case regions between the Empirical Characteristic Functions (ECFs) of the target and generated distributions. The GNN is trained to move these points toward the areas of highest discrepancy, while maintaining locality and dynamically adapting to changes in the loss surface. 

\begin{remark}
By leveraging the graph structure, the GNN helps identify the worst-case discrepancies between the empirical characteristic functions (ECFs) of real and generated data, ensuring efficient convergence to the target distribution. This process allows the generator to continuously improve its approximation of the target distribution based on the worst-case scenario identified by the GNN.
\end{remark}

The detailed steps of our GNN-based distribution optimizer are summarized in Algorithm \ref{alg:alg_gnn_final}, which outlines the iterative process of computing ECFs for both real and generated data, followed by the dynamic optimization of the GNN to identify the most critical regions of discrepancy. 

\begin{algorithm}
    \DontPrintSemicolon	
    \SetNoFillComment
    {
        \KwInput{Real data distribution $\mathcal{P}_d$; Gaussian noise $\mathcal{P}_\mathcal{N}$; batch sizes $b_d$, $b_g$, and $b_t$ for real data, generator noise, and GNN input points, respectively; learning rates $l_r^G$, $l_r^{GNN}$ for the generator and GNN}
        \KwOutput{Trained generator $G$ and GNN parameters $\bm{\theta}_{t}$}
        
        \While{training not converged}
        {
            \tcc{Step 1: Compute ECF for real data}
            Sample real data $\{\overline{\mathbf{x}}_i\}_{i=1}^{b_d}\sim\mathcal{P}_d$ and compute the ECF: 
            $\widehat{\Phi}_{\bm{\mathcal{X}}_n}(\mathbf{t}) = \frac{1}{n} \sum_{i=1}^n e^{j\mathbf{t}^T \mathbf{x}_i}$

            \tcc{Step 2: Generate data from the generator}
            Sample generator input noise $\{\mathbf{z}_i\}_{i=1}^{b_g}\sim\mathcal{P}_\mathcal{N}$ and generate data $\{\mathbf{y}_i\}_{i=1}^{b_g} = G(\{\mathbf{z}_i\}_{i=1}^{b_g})$\;

            \tcc{Step 3: Compute ECF for generated data}
            Compute the ECF of the generated data: 
            $\widehat{\Phi}_{\bm{\mathcal{Y}}_n}(\mathbf{t}) = \frac{1}{n} \sum_{i=1}^n e^{j\mathbf{t}^T \mathbf{y}_i}$

            \tcc{Step 4: Prepare query points with loss information}
            Sample Gaussian points as query points$\{\mathbf{t}_i\}_{i=1}^{b_t}\sim\mathcal{P}_\mathcal{N}$ and compute the discrepancy: 
            $L_{\{\mathbf{t}_i\}_{i=1}^{b_t}} = |\widehat{\Phi}_{\bm{\mathcal{X}}}(\mathbf{t}) - \widehat{\Phi}_{\bm{\mathcal{Y}}}(\mathbf{t})|$
            
            Augment each query point $\mathbf{t}_i$ by concatenating it with the discrepancy:
            $\mathbf{t}^{aug}_i = [\mathbf{t}_i, L(\mathbf{t}_i)]$

            \tcc{Step 5: Update the GNN}
            Compute updated query points using the GNN: 
            $\mathbf{t}^{updated}_i \leftarrow \text{GNN}(\mathbf{t}^{aug}_i)$

            Compute the new discrepancy loss using the updated points: 
            $L_{\{\mathbf{t}^{updated}_i\}_{i=1}^{b_t}} = |\widehat{\Phi}_{\bm{\mathcal{X}}}(\mathbf{t}^{updated}_i) - \widehat{\Phi}_{\bm{\mathcal{Y}}}(\mathbf{t}^{updated}_i)|$
            
            Update the GNN parameters $\bm{\theta}_{t}$:
            $\bm{\theta}_{t} \leftarrow \bm{\theta}_{t} + l_r^{GNN} \cdot \mathrm{Adam}(\nabla_{\bm{\theta}_{t}}[L_{\{\mathbf{t}^{updated}_i\}_{i=1}^{b_t}}])$

            \tcc{Step 6: Update the generator}
            After updating the GNN, forward the GNN again to compute new updated points $\mathbf{t}^{updated}_i$: 
            $\mathbf{t}^{updated}_i \leftarrow \text{GNN}(\mathbf{t}^{aug}_i)$

            Compute the generator’s loss using these new points: 
            $L_{\{\mathbf{t}^{updated}_i\}_{i=1}^{b_t}} = |\widehat{\Phi}_{\bm{\mathcal{X}}}(\mathbf{t}^{updated}_i) - \widehat{\Phi}_{\bm{\mathcal{Y}}}(\mathbf{t}^{updated}_i)|$
            
            Update the generator parameters $\bm{\theta}_g$: 
            $\bm{\theta}_g \leftarrow \bm{\theta}_g - l_r^G \cdot \mathrm{Adam}(\nabla_{\bm{\theta}_g}[L_{\{\mathbf{t}^{updated}_i\}_{i=1}^{b_t}}])$
        }
    }
    \caption{GNN-based distribution optimizer for matching ECFs.}
    \label{alg:alg_gnn_final}
\end{algorithm}

\subsection{Graph Construction and Stabilizing the Sampling Process}

Since characteristic functions (CFs) are automatically aligned around zero, the initialization with Gaussian sample points provides a reasonable starting point for optimization. This approach not only ensures that the initial values around zero are well represented, but also indirectly focuses on matching higher-order statistics, as the $n$-th order derivative of the CF at zero corresponds to the $n$-th order moment of the distribution. From these initial points, we then construct a graph whereby the nodes represent the generated sample points, and the edges between nodes are determined based on the distance between points. Closer points are more likely to be connected, with the probability of forming an edge inversely related to the distance. This graph structure is essential for preserving locality, as it allows the GNN to focus on nearby sample points during the optimization process, ensuring that the network uses only local information to update the sample points efficiently.

In order to further stabilize the sampling process, we adopt percentage-based adjustments instead of directly adding residuals to the sample points. This approach ensures that updates are scaled relative to the original coordinates, making the changes consistent, regardless of the distance from the origin. By learning a percentage change and applying it uniformly, the GNN maintains stability across all sample points, thus preventing extreme updates and ensuring a smoother convergence towards the worst-case regions of the loss surface.

\subsection{Dynamic Optimization of the Loss Surface}

The second advantage of GNNs lies in their ability to handle dynamic changes in the loss surface. As the generator is updated, the loss surface—based on the discrepancy between the ECFs of the real and generated data—shifts. This means that previously identified worst-case points may no longer be optimal as the training progresses.

To account for these changes, the GNN incorporates local information from the evolving loss surface. The network aggregates information from neighboring points within the graph to better understand the local loss landscape. By augmenting each sample point with an additional dimension representing its loss value, the GNN can incorporate both spatial and loss surface information into its update process. The sample points first act as queries into the loss surface, gathering information about the local discrepancies, which is then incorporated into the GNN. The GNN uses this combined information to output new points that correspond to regions with the most significant discrepancies.

By continuously adapting to the evolving loss surface, the GNN efficiently acts as a worst-case optimizer, shifting the query points toward the most critical areas between the real and generated distributions. This dynamic optimization capability is crucial for minimizing discrepancies as the training process evolves.

\subsection{Preliminary Validation: Proof of Concept Experiment}

To demonstrate the effectiveness of our graph-based architecture, we first conducted a simple proof-of-concept experiment which highlights the advantages of GNNs in this context over traditional methods such as Gaussian sampling and fully connected networks. In this experiment, we randomly generated 1,000 sets of parameters for both the training and testing sets, each corresponding to a different Gaussian probability density function (PDF). These PDFs serve as a proxy for the empirical characteristic function (ECF) difference surface, which represents the discrepancy between two distributions. The goal of the optimizer is then to identify the regions where the two ECFs differ the most, which is analogous to finding the regions where the total value of the sample points on the PDF is maximized.

We tested three methods for finding the worst-case regions: Gaussian sampling, fully connected neural network, and our proposed GNN-based approach. The log-likelihood of the query points was used as the evaluation metric, where a higher value indicates better identification of the regions with significant discrepancies.

\begin{table}[h]
\centering
\begin{tabular}{|c|c|c|}
\hline
\textbf{Method}             & \textbf{Mean Log-Likelihood} & \textbf{Standard Deviation} \\ \hline
GNN                         & -48                          & 135                         \\ \hline
Fully Connected Network      & -199                         & 2,098                       \\ \hline
Gaussian Sampling            & -488                         & 5,607                       \\ \hline
\end{tabular}
\caption{Comparison of Methods: Mean and Standard Deviation of Results}
\label{tab:comparison_results}
\end{table}

As shown in Table \ref{tab:comparison_results}, the GNN outperformed the other considered methods, achieving a mean log-likelihood of -48 with a standard deviation of 135. In comparison, the fully connected network achieved a mean of -199 with a standard deviation of 2,098, and Gaussian sampling resulted in a mean of -488 with a standard deviation of 5,607. These results demonstrate that the GNN is not only more effective in identifying regions of high discrepancy but also more stable, as indicated by the lower variance in its results.

This experiment, while not directly related to distribution learning, provides early evidence of the effectiveness of GNNs in navigating and optimizing dynamically changing landscapes, which is crucial for optimizing the ECF loss.

\section{Experiments}

We next evaluated the ability of our approach to convert non-generative into generative models by comparing the feature space of a pre-trained autoencoder using the CelebA dataset \cite{liu2015deep} as the training dataset. We then compared the effectiveness of the optimizer across three configurations: Gaussian-based sampling, fully connected network, and our Graph Neural Network (GNN) model.

\subsection{Learning the Feature Space of a Pre-Trained Autoencoder}

The first step in our experiment was to train an autoencoder on the CelebA dataset, which then serves as the target distribution that our model aims to learn. The CelebA dataset is composed of over 200,000 celebrity images, providing a rich source of facial data for training. After training, the autoencoder produces a compressed latent space, which we treat as the target feature space for this experiment.

\begin{remark}
    Importantly, as shown in section \ref{sec:Learning_with_the_Proposed_Model} after the autoencoder is trained, we do not need to fine-tune it further. Our proposed model directly learns the distribution of the pre-trained feature space without requiring additional updates to the autoencoder itself. This capability makes our approach highly transferable, as it can turn any pre-trained, non-generative model into a generative one, enabling it to produce samples from the learned distribution. This flexibility is a key advantage, allowing generative modeling without modifying the original model.
\end{remark}

\begin{figure}[H]
    \centering
    \includegraphics[width=\linewidth]{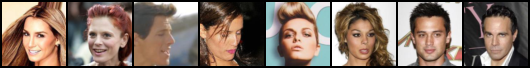}
    \caption{Output samples from the pre-trained autoencoder, serving as the target distribution.}
    \label{fig:original_samples}
\end{figure}

Figure \ref{fig:original_samples} shows the output samples from the pre-trained autoencoder, which will serve as the target distribution.

\subsection{Learning with the Proposed Model}
\label{sec:Learning_with_the_Proposed_Model}

\begin{figure}[H]
    \centering
    \includegraphics[width=\linewidth]{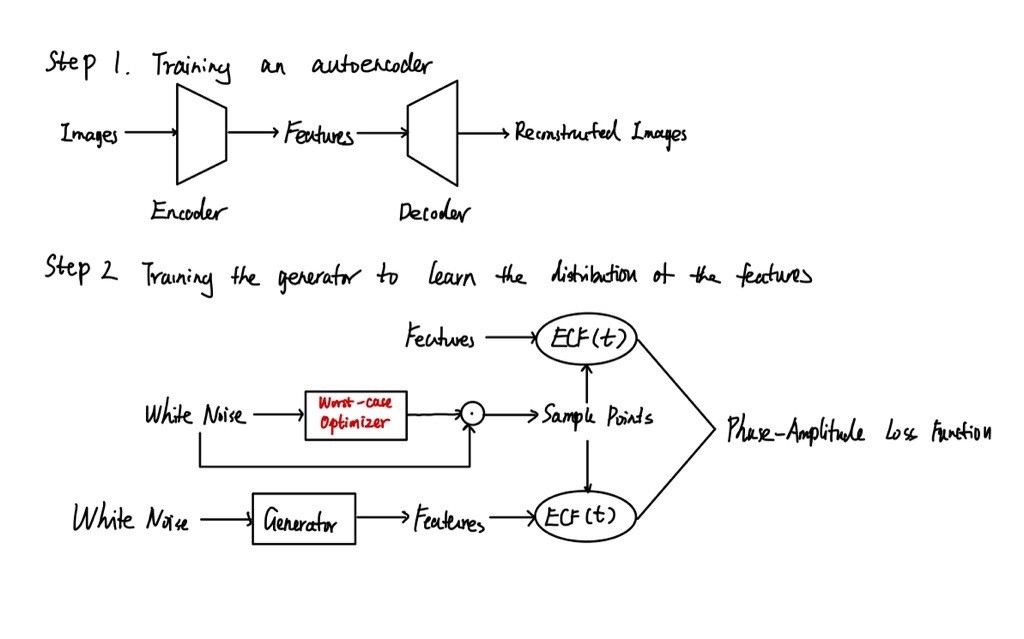}
    \caption{The proposed model for learning the feature space of a pre-trained autoencoder. The generator produces data from Gaussian noise, and the optimizer (Gaussian-based, fully connected, or GNN) identifies regions of greatest discrepancy in the feature space.}
    \label{fig:ae_structure}
\end{figure}

After training the autoencoder, we proceed with learning the target distribution using our model. The generator in our model takes Gaussian noise as input and aims to generate data that matches the feature space of the pre-trained autoencoder. The optimizer, as shown in Figure \ref{fig:ae_structure}, serves as the worst-case optimizer, plays a crucial role in identifying discrepancies between the real and generated data distributions.

We explored three variations of the optimizer in this experiment:
\begin{enumerate}
    \item \textbf{Gaussian-based method:} This is similar to our previous work \cite{9730054}, where we employed Gaussian sampling to define the sample points for the loss function.
    \item \textbf{Fully connected network:} A fully connected network was used as the optimizer, learning to identify worst-case regions in the feature space.
    \item \textbf{Graph Neural Network (GNN):} Our proposed GNN-based approach, which incorporates locality and dynamically optimizes the loss surface.
\end{enumerate}

Figure \ref{fig:ae_structure} illustrates the structure of our model, showing the role of the generator, optimizer, and the loss surface optimization.

\subsection{Comparison of Worst-Case Optimizer Configurations}

We compared the generated samples for each of the three configurations: Gaussian-based sampling, fully connected network, and the proposed GNN-based optimizer. In each case, we evaluated the quality of the samples produced and the ability of the optimizers to identify the most significant regions of the loss surface.

\begin{figure}[htbp]
    \centering
    \subfigure[Samples generated using Gaussian-based sampling.]{
        \includegraphics[width=\linewidth]{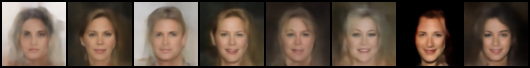}
        \label{fig:gaussian_samples}
    }
    
    \vspace{0.1cm} 

    \subfigure[Samples generated using fully connected network-based optimizer.]{
        \includegraphics[width=\linewidth]{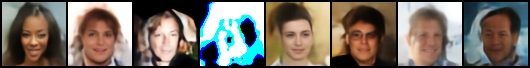}
        \label{fig:fcn_samples}
    }
    
    \vspace{0.1cm} 

    \subfigure[Samples generated using GNN-based optimizer.]{
        \includegraphics[width=\linewidth]{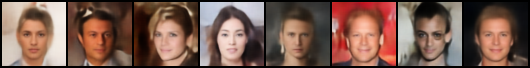}
        \label{fig:gnn_samples}
    }

    \caption{Generated samples using different optimization methods: (a) Gaussian-based sampling, (b) Fully connected network-based optimizer, (c) GNN-based optimizer.}
    \label{fig:combined_samples}
\end{figure}

Figure \ref{fig:gaussian_samples} shows the generated samples using Gaussian-based sampling. These samples, while capturing some aspects of the target distribution, exhibit a lack of diversity and tend to converge towards similar modes. Figure \ref{fig:fcn_samples} illustrates the results using a fully connected network as the optimizer, which the generated samples showing enhanced variety, but certain features of the target distribution are not well captured. Finally, Figure \ref{fig:gnn_samples} presents the results using our GNN-based optimizer. The samples generated using the GNN exhibit the highest fidelity to the target distribution, with clear diversity and sharper features. This demonstrates the ability of GNNs to incorporate locality of the information and adapt dynamically to the changing loss surface; thus allows them to capture a broader range of features from the target distribution.

\begin{remark}
    Figure \ref{fig:combined_samples} demonstrates the clear advantage of the GNN-based optimizer in terms of sample quality and diversity. Through dynamical optimization and by leveraging local information in the feature space, the GNN model successfully identifies critical regions, leading to improved performance over Gaussian sampling and fully connected networks.
\end{remark}

\section{Conclusion}

We have introduced a graph neural network (GNN)-based approach for optimizing the discrepancy between empirical characteristic functions (ECF) of real and generated data. By leveraging the locality-preserving properties of GNNs and their ability to dynamically adapt to changes in the loss surface, our method has been shown to effectively identify the worst-case regions in the distribution space, allowing for efficient training of the generator. We have demonstrated the advantages of our approach over Gaussian-based methods and fully connected networks in both a preliminary validation experiment and a more complicated scenario involving the feature space of a trained autoencoder on the CelebA dataset. The experiments have demonstrated that the GNN-based optimizer significantly outperformed traditional sampling methods, particularly in cases where the underlying data distribution is complicated. The locality and dynamic optimization capabilities of the GNN have proven to be crucial in improving the identification of critical regions of discrepancy between distributions, resulting in better alignment between the real and generated data. Our proposed method has also been shown to offer a scalable and efficient way to convert pre-trained, non-generative models into generative ones, demonstrating its broader applicability beyond the specific case of autoencoder feature space learning. Future work may explore further optimizations of the graph neural network architecture and applications to other tasks involving high-dimensional distribution matching.

\bibliographystyle{splncs04}
\bibliography{ShengxiLi}

\end{document}